\documentclass[journal]{IEEEtran}
\usepackage{amsmath,amsfonts}
\usepackage{array}
\usepackage[caption=false,font=normalsize,labelfont=sf,textfont=sf]{subfig}
\usepackage{textcomp}
\usepackage{ulem}
\usepackage{stfloats}
\usepackage{url}
\usepackage{verbatim}
\usepackage{graphicx}
\usepackage{cite}
\usepackage{graphicx}
\usepackage{float}
\usepackage{xspace}
\usepackage{amsmath}
\usepackage{graphicx}
\usepackage{multirow}
\usepackage{booktabs}
\usepackage{widetable}
\usepackage{adjustbox}
\usepackage{makecell}
\usepackage{color, colortbl}
\usepackage{xcolor}
\usepackage{amssymb}
\usepackage{algorithm}
\usepackage{algorithmic}
\usepackage{hyperref}

\newcommand{\figref}[1]{Fig.~\ref{#1}}
\newcommand{\tabref}[1]{Table~\ref{#1}}

\def\onedot{\ifx\@let@token.\else.\null\fi\xspace}

\definecolor{amber}{rgb}{1.0, 0.75, 0.0}
\definecolor{mygray}{gray}{.9}

\makeatletter
\newcommand{\thickhline}{
	\noalign {\ifnum 0=`}\fi \hrule height 1pt
	\futurelet \reserved@a \@xhline
}
\makeatother

\begin{document}

\title{TransBridge: Boost 3D Object Detection by Scene-Level Completion with Transformer Decoder}

\author{Qinghao Meng,
        Chenming Wu,
        Liangjun Zhang,
        and Jianbing Shen,
        
\IEEEcompsocitemizethanks{
\IEEEcompsocthanksitem Q. Meng is with the School of Computer Science, Beijing Institute of Technology, Beijing, China.
\IEEEcompsocthanksitem C. Wu and L. Zhang are with the Robotics and Autonomous Driving Lab (RAL), Baidu Research, China. C. Wu is the project lead. 
\IEEEcompsocthanksitem J. Shen is with the State Key Laboratory of Internet of Things for Smart City, Department of Computer and Information Science, University of Macau, Macau, China.
(Email: jianbingshen@um.edu.mo)

\IEEEcompsocthanksitem  Corresponding author: \textit{Jianbing Shen}.
}
\thanks{}
}

\markboth{}
{Meng \MakeLowercase{\textit{et al.}}}

\maketitle

\begin{figure*}[h]
\centering
\includegraphics[width=0.99\linewidth]{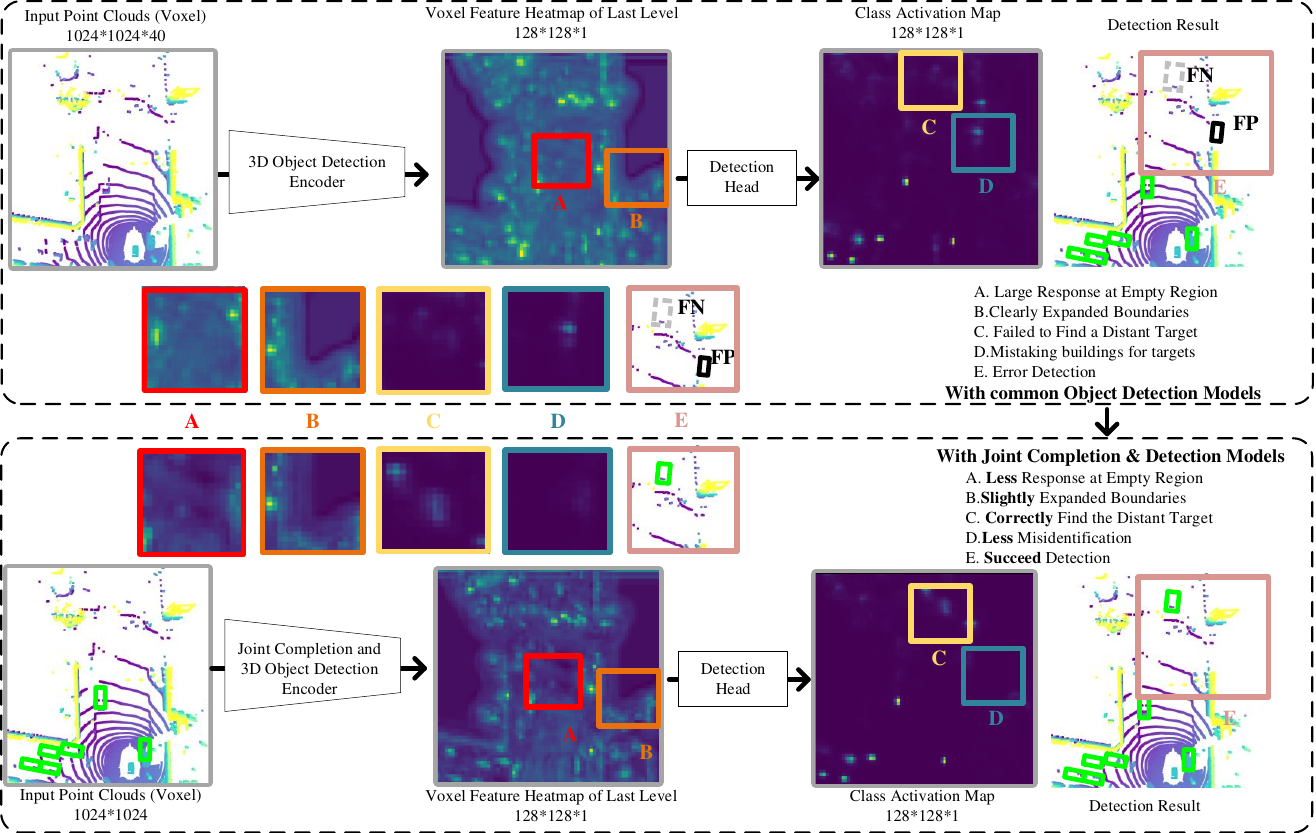}
\centering
\caption{\textbf{LiDAR-based 3D object detection performed without (top) and with (bottom) point cloud completion.} {In this figure, we illustrate a portion of the input point cloud along with the corresponding feature map for this region.}  
The basic 3D Object Detection Encoder is not trained to distinguish between empty regions that are truly invisible voxels or transparent voxels.
Through jointly learning 3D object detection and point cloud completion, feature maps and class maps become clearer, resulting in improved detection results.
}
\label{fig:compare}
\end{figure*}

\begin{abstract}
3D object detection is essential in autonomous driving, providing vital information about moving objects and obstacles. 
Detecting objects in distant regions with only a few LiDAR points is still a challenge, and numerous strategies have been developed to address point cloud sparsity through densification.
This paper presents a joint completion and detection framework that improves the detection feature in sparse areas while maintaining costs unchanged. Specifically, we propose \textit{TransBridge}, a novel transformer-based up-sampling block that fuses the features from the detection and completion networks.
The detection network can benefit from acquiring implicit completion features derived from the completion network. Additionally, we design the \textit{Dynamic-Static Reconstruction} (DSRecon) module to produce dense LiDAR data for the completion network, meeting the requirement for dense point cloud ground truth.
Furthermore, we employ the transformer mechanism to establish connections between channels and spatial relations, resulting in a high-resolution feature map used for completion purposes.
Extensive experiments on the nuScenes and Waymo datasets demonstrate the effectiveness of the proposed framework. 
The results show that our framework consistently improves end-to-end 3D object detection, with the mean average precision (mAP) ranging from 0.7 to 1.5 across multiple methods, indicating its generalization ability. For the two-stage detection framework, it also boosts the mAP up to 5.78 points.  
\end{abstract}


\section{Introduction}
\label{sec:intro}
\IEEEPARstart{T}{hanks} to the high precision in sensing distances in surrounding environments, 3D object detection technology has become pivotal in the field of autonomous driving~\cite{arnold2019survey}. 
In this context, LiDAR-based 3D object detection plays an important role in making driving decisions and assisting self-driving vehicles. There have been extensive research efforts into the development of this field. 
For instance, VoxelNet~\cite{zhou2018voxelnet},
  CenterPoint~\cite{yin2021center}, and VoxelNext~\cite{chen2023voxelnext} and many well-labeled datasets (e.g., KITTI~\cite{geiger2012we}, nuScenes~\cite{nuscenes2019}, 
  and Waymo~\cite{sun2020scalability}) continuously push the boundaries of this task.

In previous work, a primary challenge in LiDAR-based 3D object detection is the inherent sparsity and nonuniformity of the LiDAR data~\cite{li2021improved,shi2023real,ning2022point}. {Generating representative features from these regions remains challenging.}
When transforming the point cloud into voxels for feature extraction, the empty regions become empty voxels, making it difficult to distinguish whether obstacles exist inside them.
We categorize these voxels into two types: ``transparent'' voxels, which are empty due to the absence of obstacles, and ``invisible'' voxels, which result from occlusions or insufficient LiDAR resolution. 
When we feed these empty voxels to convolutional networks,
the produced feature maps tend to cover a larger area than that occupied by objects, overlooking invisible voxels. 
This discrepancy results in both false negatives (FN) and false positives (FP), as shown in~\figref{fig:compare}. Consequently, exploring ways to minimize the influence of invisible voxels is essential.
 
To address this issue, point cloud completion networks~\cite{yuan2018pcn,yang2018foldingnet,yi2021complete,zhang2022probabilistic} are adapted to 3D object detection frameworks.
For example, techniques like PC-RGNN~\cite{zhang2021pc} have been employed, leveraging completion modules trained with different datasets like ShapeNet~\cite{chang2015shapenet}, which provide dense synthetic point clouds of cars.
Some approaches like SPG~\cite{xu2021spg} and PG-RCNN~\cite{koo2023pg} proposed completion modules trained with LiDAR point clouds.
Some approaches~\cite{qi2021offboard,koh2022mgtanet} have also integrated consecutive LiDAR frames to densify point clouds, while others utilized a two-stage detection process to densify the point cloud with temporal data.
Some methods also employ a two-stage detection approach. In the second stage, the feature of non-empty voxels is diffused into empty voxels, creating a dense feature map to improve detection effectiveness, such as PV-RCNN~\cite{shi2020pv}.
While numerous strategies have been developed to address point cloud sparsity in 3D object detection through densification, these approaches significantly increase computational demands during inference. Furthermore, they have not cultivated a more robust feature extractor capable of differentiating empty voxels. 
Integrating detection and completion learning within a single neural network is essential to maintain the original inference speed while enabling the detection of invisible voxels.

In this paper, we proposed a shared pyramid network as the encoder and created a unique decoder for the completion task. Specifically, the detection network typically employs a pyramid structure, while the completion network consists of a pyramid and an inverted pyramid. The extra supervision from point cloud completion makes the feature maps clearer and improves detection results. 
As shown in~\figref{fig:compare}, we have extracted a portion of the point cloud for visualization, illustrating the feature extraction and object detection process within this region.
To better represent the feature maps, we combine the features from all channels and normalize them, where yellow represents higher values, blue represents lower values, and greater color contrast indicates larger feature gaps between adjacent regions.
The detector takes the point cloud data as input, and produces detection results after feature extraction. We visualize the feature maps from the last convolution layer before the detection head (A-C) and the class activation maps for vehicle targets (D). 
The green regions with medium values in the feature maps occupy most of the area, indicating the detector's limited ability to distinguish between obstacles and backgrounds.
These responses introduce interference to object detection: for instance, the location of vehicle C, which contains a sparse number of features and a high proportion of invisible voxels, is overwhelmed by noise. Furthermore,  a wall segment in D is misidentified as a vehicle target in the class activation map due to significant noise. As a result, the detection performance suffers and causes false negatives (FN) in C-E and false positives (FP) in D-E.

Importantly, although the entire network is jointly trained, object detection results are obtained immediately after the feature encoder's calculation, enabling optimal point cloud completion costs during inference.

The completion decoder performs feature up-sampling at each level. Each decoder layer incorporates a completion head to predict the visibility status of voxels.
While learning to distinguish between invisible and transparent voxels, the completion decoder guides the shared encoder to generate features with rich structural information, particularly in scenarios where LiDAR data is sparse or partially occluded.
However, incorporating detection and completion training in a single network presents challenges due to their vastly different feature distributions.
Point cloud completion typically operates on sparse input data, aiming to fill in missing regions in the point cloud. This task requires the network to focus on local geometric structures and patterns by irregular and varying densities. 
In contrast, 3D object detection involves identifying and classifying objects of interest, focusing on global features such as object shape, size, and orientation~\cite{yin2021center}.
On the other hand, detection networks heavily rely on semantic features that indicate object categories and their contextual relationships within the environment~\cite{zhang2023few}. Due to these differences in feature emphasis, using identical network layers or modules may not be equally effective for both tasks. Simply cutting channels in the network might lead to training failure~\cite{yi2021complete}.

To address this challenge, we introduce \textit{TransBridge}. This component serves as both an upsampling module and a mediator between completion and detection features, incorporating a transformation mechanism~\cite{pan20213d}.
The TransBridge block processes detection features as input and produces completion features as output.
Furthermore, a \textit{sparsity control module} (SCM) is incorporated to identify voxels, controlling the number of voxels involved in the calculations.
The training of TransBridge relies on dense point clouds as ground truth.
A straightforward approach is to merge point cloud sequences~\cite{yan2021sparse}, which may create noise such as trailing smear after moving targets.
To obtain clean point cloud data, we follow methods in~\cite{zheng2022boosting,wang2023implicit} to separate and align dynamic objects to their coordinates.
However, we observe that sequence point clouds remain sparse in distant regions.
To resolve this, we implement surface reconstruction NKSR~\cite{huang2023neural} after split alignment, a process we term \textit{Dynamic-Static Reconstruction} (DSRecon).
{The proposed TransBridge module has profound implications for intelligent transportation, specifically in improving the safety and efficiency of self-driving systems. The joint completion and detection framework improves the precision and reliability of the perception system, such as detecting distant cars at intersections, which is vital for collision avoidance and path planning. 
Importantly, our method does not introduce additional computational overhead during runtime, ensuring the efficiency of the system. Our approach optimizes detection accuracy while maintaining real-time capabilities, making it a valuable contribution to the field of intelligent transport systems.}
Our main contributions are summarized below:

\begin{itemize} \setlength{\itemindent}{-0.1cm}
\item We propose a novel \textit{Detection and Completion Framework}.
In this framework, LiDAR-based 3D object detection and scene-level completion networks are jointly trained end-to-end.
While both networks utilize the same feature encoder, they have separate output modules. Detection is performed through the detection head and completion is accomplished through the completion decoder.
This shared encoder is adaptable to various detection encoders, enabling easy integration with other detectors.
\item We design \textit{TransBridge block}, a transformer-based point cloud decoder.
It includes an Up-Sampling Bridge for extracting informative features and an Interpreting Bridge to facilitate efficient feature transfer between the detection and completion networks.
 We equip our completion network with TransBridge blocks to compensate for information loss during upsampling and interpret the distinct semantic meanings in detection and completion features. 
\item We introduce the \textit{Dynamic-Static Reconstruction} (DSRecon) Module.
It generates high-fidelity, dense ground truth for learning point cloud completion, where we merge the point cloud sequences and align dynamic foregrounds to their coordinates, thereby reducing trailing smear effects.
The point cloud is subsequently enhanced using surface reconstruction methods.

\item We present the \textit{Sparsity Controlling Module} (SCM).
This module ensures efficient computation by regulating the number of non-empty voxels and eliminating transparent voxel features generated during the upsampling process.
\end{itemize}

\section{Related Work}
\subsection{LiDAR-based 3D object detection} 
{LiDAR sensors effectively capture depth information in 3D space. Through analysis of LiDAR data, we significantly enhance the environmental understanding~\cite{liu2022cross,lyu2023self}. 
This capability is crucial for intelligent transportation systems, where accurate environmental perception improves safety, efficiency, and reliability.}
Recently, researchers focused extensively on detecting 3D objects from LiDAR point clouds using voxel-based encoders~\cite{
liu2020tanet,shi2022pillarnet,
wang2024club,li2023pillarnext}.
Point-based detectors have also been developed~\cite{chen2022sasa,
ahmed2022smart}.
Inspired by the RGB-D based detection algorithm~\cite{sun2023saliency}, \cite{tian2022fully} and \cite{bai2024rangeperception} process the LiDAR point cloud as a range-image, and ~\cite{zhou2023octr} employs an Octree-based network.
~\cite{zhang2022dsp} optimizes object detection systems with Internet of Things (IoT)-based hardware.
Subsequent research made significant contributions to network design~\cite{chen2022focal,sheng2022rethinking,hu2023collaboration}.
In addition, ~\cite{yin2022semi,xu2022back,zhao2022prototypical,yin2022proposalcontrast} propose semi-supervised and unsupervised learning frameworks, and ~\cite{lu2023open,pang2023unsupervised} learn 3D object detection without 3D annotation.

Various methods can enhance LiDAR-based 3D object detection: \textit{a) With Virtual Point Clouds:} Studies like \cite{li2021sienet} synthesize the 3D models of cars to align detection proposals. Recent approaches \cite{shin2023diversified} extract foreground objects from real-world data. \cite{qi2021offboard, koh2023mgtanet, nie2023glt} employ entire sequences of multiple LiDAR sweeps to enhance temporal accuracy.

\textit{b) With Depth:} Depth completion networks use images to determine depth information. This information is then converted into a pseudo point cloud for detection, as demonstrated in SPD \cite{wu2022sparse} and MVP \cite{yin2021multimodal}.
\textit{c) With Distillation:} SMF-SSD~\cite{zheng2022boosting} presents a learning framework based on a distillation network. The student network is designed to mimic the features of a teacher network trained on synthesized dense point clouds. The student's performance is constrained by the teacher's features. 
Unlike existing methods, we enhance the detection network through an additional decoder network which performs joint point cloud completion learning with 3D object detection.

\begin{figure*}
\includegraphics[width=\linewidth]{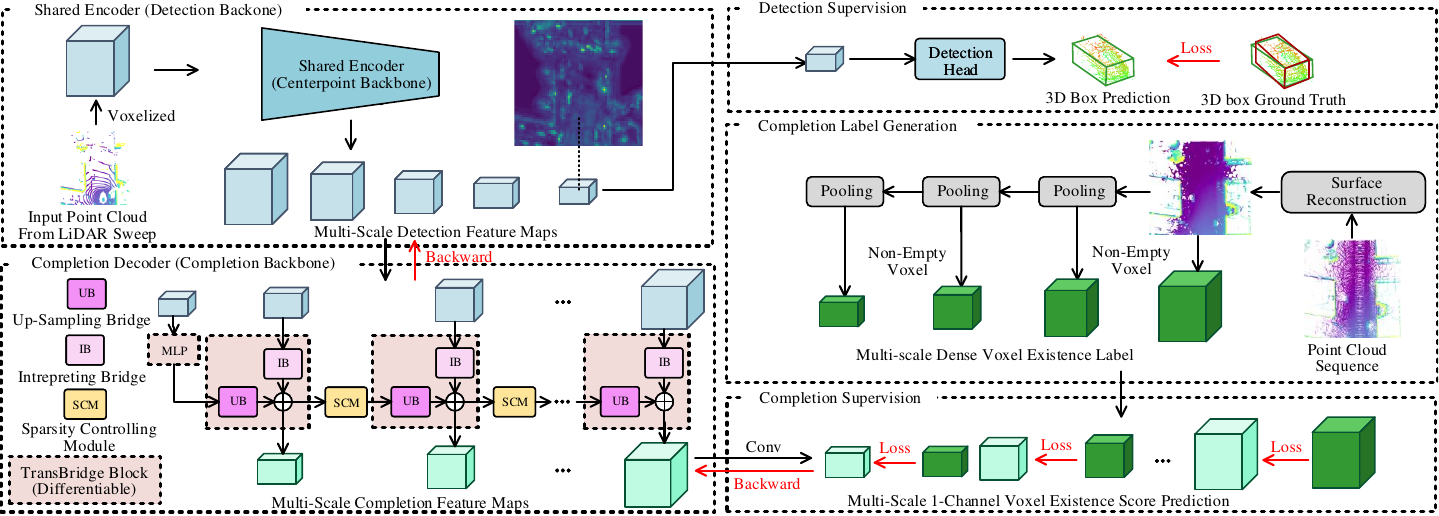}
\centering
\caption{\textbf{The TransBridge detection-completion pipeline.}
We use the Centerpoint~\cite{yin2021center} as an example.
which is shown in \textit{Shared Encoder}. Then, the detection output is supervised with 3D bounding box groundtruth.
In \textit{Completion Decoder}, the TransBridge blocks process multi-scale detection features maps and generate voxel existence scores.
Each TransBridge block contains an Up-Sampling Bridge (UB) and an Interpreting Bridge (IB). Features from these two modules are concatenated and processed with the Sparsity Controlling Module (SCM).
Finally, surface reconstruction is conducted on merged  point cloud sequences to generate a dense point cloud and perform voxelization to obtain the dense voxel existence label as the supervision of the completion decoder.
}
\label{fig:framework}
\end{figure*}

\subsection{Point Cloud Completion}
Completing 3D shapes by point cloud completion has emerged as a significant research area~\cite{kazhdan2013screened}.
Recent deep neural networks focus on indoor scenarios \cite{cao2023kt, zhang2023few, zhang2022probabilistic, weinzaepfel2022croco, cheng2023sdfusion}, primarily utilizing RGBD cameras \cite{wu2023omniobject3d} or synthetic data~\cite{lamb2023fantastic}.
Completing 3D object shapes commonly involve partial scans \cite{yang2018foldingnet, yu2022point, zhou2023sparsefusion}, implementing attention blocks similar to those in SA-Net \cite{wen2020point} for synthesis point cloud completion. PMP-Net \cite{wen2021pmp} innovatively moves redundant points to restore missing parts.
In outdoor scenes, recent methods such as \cite{wang2023autorecon} address point cloud completion, especially in autonomous driving scenarios \cite{zhang2021pc, weber2023power}.
These methods train their networks on synthesis datasets and evaluate on real LiDAR data.  In contrast, studies like \cite{xu2021spg, xu2022behind} utilize point cloud data of stationary cars from actual LiDAR datasets as training data.
\cite{ye2023LiDARmultinet} simultaneously optimizes both the detection and segmentation heads.
This dual optimization enhances better detection results.
However, the incorporation of completion and refinement stages increases computational time.

A common technique for obtaining ground truth for point cloud completion in outdoor scenes involves merging sequential data. However, dynamic targets such as vehicles and pedestrians invariably create trailing smear effects in the merged point clouds.
To address the completion of point clouds without trailing smear, JS3C-Net \cite{yan2021sparse} and PG-RCNN \cite{koo2023pg} utilize only foreground instance point clouds from consecutive frames in autonomous driving datasets, such as nuScenes \cite{nuscenes2019} and Waymo \cite{sun2020scalability}.  
These systems acquire point cloud data from each LiDAR sweep and develop 3D point cloud completion models by reconstructing dense point clouds within the detection proposals.
Additionally, approaches such as SVCN \cite{yi2021complete} combine point cloud sequences and implement Poisson surface reconstruction and surface trimming techniques. This enhances the density of the background point clouds and aids in the training for domain adaptive segmentation. 
In our approach, we independently collect dynamic foreground and static background point clouds~\cite{koo2023pg}. Moreover, we perform surface reconstruction as demonstrated in~\cite{yi2021complete} on both data sources and integrate them to generate clear and dense point clouds for scene-level completion training.

\section{Proposed Method}
\label{sec:method}

\subsection{Overview}
\label{subsec:all}

A pipeline for LiDAR-based 3D object detection and point cloud completion is illustrated in \figref{fig:framework}. By integrating voxel-based Centerpoint \cite{yin2021center} 3D object detection with our novel TransBridge Completion backbone, we demonstrate the effectiveness of joint learning detection and completion. The detection encoder (in {\color{blue}blue}) generates multi-scale detection feature maps and produces 3D bounding box predictions.
Then, the completion decoder (in {\textcolor[RGB]{116,201,120}{green}}) takes the detection feature maps as input and generates completion feature maps to represent voxel existence, where the transparent voxels indicate non-existing and invisible voxels.
Each TransBridge block comprises an Up-sampling Bridge (UB) to enhance the resolution of feature maps and an Interpreting Bridge (IB)  to interpret detection-to-completion semantics.
Additionally, Sparsity Controlling Modules (SCMs) are implemented after channel-wise concatenation, preserving the sparsity of high-resolution feature maps.
The training process utilizes 3D bounding box annotations to learn 3D object detection and employs dense point clouds from the DSRecon module to complete point clouds.
The TransBridge blocks, being differentiable, enable gradient back-propagation from the completion to the detection encoder.

For clarity, we denote the detection feature map at resolution level $i$ as $f_D^i$. Following \cite{yan2018second}, we voxelize the point clouds to obtain a 3D feature map of the first level, $f_{input}=f_D^1$.
Detection blocks ($\mathbf{D}^i$) then apply 3D sparse convolutions and pooling to extract features for subsequent levels:
\begin{equation}
\begin{aligned}
f_D^{i+1} = \mathbf{D}^i(f_D^{i}),~\text{for}~i\in \{1,2...N-1\}~~~~~~~~
\end{aligned}
\label{eq:detection}
\end{equation}
At each level, we refer to the feature map produced by the TransBridge block ($\mathbf{T}$) as $f_T^i$. This block combines the detection feature map $f_D^i$ and previous features $f_T^{i+1}$ to obtain the completion feature $f_T^{i}$:
\begin{equation}
\begin{aligned}
f_T^{i} = \mathbf{T}^i(f_D^{i},f_T^{i+1}),~\text{for}~i\in \{1,2...N-1\}~~~~~~~~
\end{aligned}
\label{eq:completion}
\end{equation}

\begin{figure*}[t!]
\includegraphics[width=1.0\linewidth]
{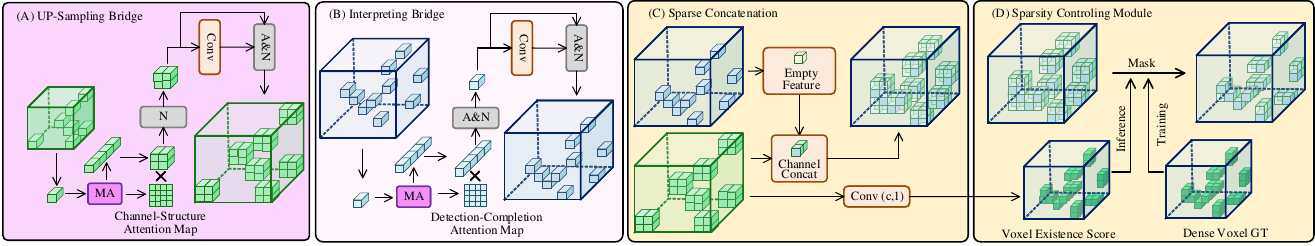}
\centering
\caption{\textbf{The detailed structure of TransBridge.} The Up-Sampling Bridge (A) takes the completion feature maps from the previous level and generates features for sub-voxels through channel-structure attention. The Interpreting Bridge (B) receives the detection feature maps and modifies the semantics of voxel features through detection-completion attention. Then, in the Sparsity Concatenation (C), since the distributions of detection voxels and completion voxels differ, thus we add empty voxels to the detection feature map. Finally, in the Sparsity Controlling Module (D), both features are concatenated to formulate new completion feature maps, and the empty voxels are removed during inference to ensure efficient training.
}
\label{fig:transbridge}
\end{figure*}

\subsection{{TransBridge Block}}
\label{sec:tranbridge}
The TransBridge block is primarily designed to up-sample the completion feature map and formulate informative completion features from detection features. The Up-Sampling Bridge block manages resolution enhancement, and the Interpreting Bridge block handles translating the detection features. Both blocks incorporate transformer structures. We describe the pooling process from $(d,w,h)$ of $c_1$ channel features to $(2d,2w,h)$ of $c_2$ channel features, where each voxel at a lower resolution level is divided into 4 sub-voxels.
To better illustrate the feature interaction process in the UB/IB block, we provide the pseudocode for the TransBridge block in \figref{fig:framework} within Algorithm.~\ref{alg:transbridge}.

\begin{figure}[t!]
\includegraphics[width=1.0\linewidth]
{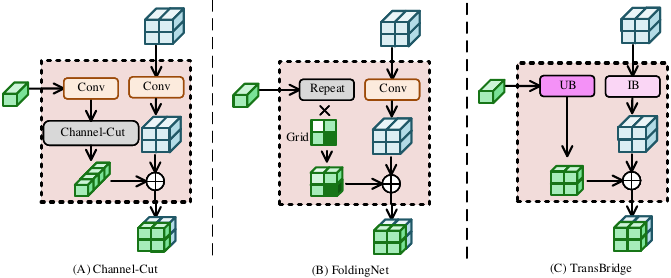}
\centering
\caption{\textbf{The comparison between different decoder networks.} (A) The Channel-Cut method conducts convolution and splits the voxel features along the channel. (B) The FoldingNet method repeats the voxel features and multiplies the features with a spatial bias to obtain sub-voxel features. (C) The TransBridge up-samples the completion features using the UB block and processes the detection features with IB block. Subsequently, all methods perform concatenation on feature maps from two sources.}
\label{fig:network}
\end{figure}

\textbf{Up-sampling Bridges.} 
A common approach to enhance the resolution of feature maps involves channel-cutting, where, for instance, input voxel features $c_1$ are processed through a Multi-Layer Perception (MLP), generating $4c_2$ channel features, which are then subdivided into four $c_2$-channel sub-voxel features, see~\figref{fig:network} (A).
 However, this technique assumes that sub-voxel features are systematically arranged in the channel sequence.
 Since detection and completion features have distinct semantics, this method can result in significant information loss.
 Another approach to upsampling features is FoldingNet~\cite{yang2018foldingnet}, which duplicates input features and multiplies features with the $(x,y,z)$ spatial coordinates. Although the information from all channels is transmitted to the next level, the interaction between voxels is limited and the parameters cannot be learned.
 To address these limitations, we introduce the transformer-based block, which creates connections between feature channels and sub-voxels, thus preserving crucial information during the up-sampling phase ($\mathbf{U}$). 

As shown in~\figref{fig:transbridge}, it processes the completion feature map from the previous level $f_T^{i+1}$ as input and obtains $f_{U}^i$. 
For each voxel's features, the block performs multi-head attention and generates a channel-structure attention feature map, which has the dimensions of $4 \times c_2$ of each head.
Then, we multiply voxel features with attention map to obtain $c_2$ features. 
We establish the one-to-one correspondence between each head and sub-voxel, transforming the $c_2$ channel features into sub-voxel features. 
Finally, we apply normalizations and convolutions to the features of each sub-voxel:
\begin{equation}
\label{eq:attention}
\begin{aligned}
f_{U}^i =& \mathbf{U}(f_T^{i+1})\\
\end{aligned}
\end{equation}
\textbf{Interpreting Bridges.} 
 The primary function of Interpreting Bridges ($\mathbf{I}$) is to interpret these detection features into completion features.
Taking the $c_2$ channel detection feature input, it performs similar multi-head attention processing and produces a detection-completion attention map, which has dimensions $c_2\times c_2$. Following the standard transformer layer, we sum up the features and conduct normalizations and convolutions on the features of each sub-voxel:
\begin{equation}
\label{eq:I}
\begin{aligned}
f_{I}^i =& \mathbf{I}(f_D^{i})\\
\end{aligned}
\end{equation}
The UB creates more new voxels in $f_{U}^{i}$ compared to the IB $f_{I}^i$.
We incorporate an empty voxel into $f_{I}^i$ to effectively merge these feature maps. 
This inclusion, as depicted in \figref{fig:transbridge} (C), enables channel-wise concatenation, generating a new completion feature map, $f_T^i$:
\begin{equation}
\label{eq:cat}
\begin{aligned}
f_T^i =f_{U}^i\oplus f_{I}^i
\end{aligned}
\end{equation}
When $i=N$, we only adopt an Interpreting Bridge block ($\mathbf{I}$) to extract  completion features $f_T^{N}$:
\begin{equation}
\begin{aligned}
f_T^{N}= \mathbf{I}(f_D^{N})
\end{aligned}
\label{eq:completion_final}
\end{equation}

\begin{algorithm}[h!]
\caption{TransBridge Block Pseudo-code}
\label{alg:transbridge}
\begin{algorithmic}[1]
\STATE\textbf{Input:} Detection feature map \(f_D\) (Eq.\eqref{eq:detection}), previous completion feature map \(f_T^{prev}\) (Eq.\eqref{eq:completion})
\\\textbf{Output:} Completion feature map \(f_T\)
\STATE\textbf{Modules:}
\\\(\text{ub\_attn} \leftarrow \text{MultiHeadAttention}(input=(q=4,c=c_2),output=(q=4,c=c_2))\)
\\\(\text{ib\_attn} \leftarrow \text{MultiHeadAttention}(input=(q=1,c=c_2),sum~output=(q=1,c=c_2))\)
\STATE\textbf{Process Up-Sampling Bridge (Eq.\eqref{eq:attention}):}
\STATE \(~{f'}_T^{prev} \leftarrow \text{MLP}{~f_T^{prev}}\) \COMMENT{$[w,d,4c_2]$}
\STATE \(attn_{ub} \leftarrow \text{ub\_attn}({f'}_T^{prev})\) \COMMENT{$[w,d,4,c_2]$}
\STATE \(\text{reshape~}attn_{ub}\) \COMMENT{$[2w,2d,c_2]$}
\STATE \(f_U \leftarrow \text{MLP}(attn_{ub})+attn_{ub}\) \COMMENT{$[2w,2d,c_2]$}
\STATE\textbf{Process Interpreting Bridge (Eq.\eqref{eq:I}):}
\STATE \(attn_{ib} \leftarrow \text{ib\_attn}(f_D)\) \COMMENT{$[2w,2d,1,c_2]$}
\STATE \(\text{reshape~}attn_{ib}\) \COMMENT{$[2w,2d,c_2]$}
\STATE \(f_I’ \leftarrow \text{ib\_conv}(attn_{ib})+attn_{ib}\) \COMMENT{$[2w,2d,c_2]$}
\STATE\textbf{Concatenate features (Eq.\eqref{eq:cat}):}
\STATE \(f_I \leftarrow \text{Add zero tensor}(f_I’)\)
\STATE \(f_T' \leftarrow \text{concat}([f_U, f_I])\) \COMMENT{$[2w,2d,2c_2]$}
\STATE \(f_T \leftarrow \text{MLP}(f_T')\) \COMMENT{$[2w,2d,c_2]$}
\STATE \textbf{return} \(f_T\)
\end{algorithmic}
\end{algorithm}

\subsection{{Sparsity Controlling Module (SCM)}}
In our approach, the sequentially placed TransBridge blocks exponentially increase the number of voxels, thereby increasing the density of feature maps and costing substantial computational resources. To address this, we introduce the Sparsity Controlling Module (SCM), which regulates the number of voxels passed to subsequent levels. This regulation is implemented through sigmoid linear layers after-$f_T^i$, producing $1$-channel voxel existence scores, $e^i$:
\begin{equation}
\label{eq:exist}
\begin{aligned}
    e^i=\textbf{Sigmoid}(\textbf{Linear}(f_T^{i}))~~~~~~~~~~~~~\\
\end{aligned}
\end{equation}

These scores differentiate transparent voxels (close to $1$) from transparent voxels (close to $0$).
For training, we employ multiscale dense point clouds that have been voxelized into $\hat{e}^i$. When working with TransBridge blocks, sequential operation is necessary to ensure proper sub-voxel training.
We apply a mask to $f_T^i$ to facilitate training at all levels using $\hat{e^i}$ derived from the dense point cloud ground truth. During inference, we use voxels with estimated scores higher than $\beta=0.7$:
\begin{equation}
\begin{aligned}
\overline{f_T^i}= \left\{
\begin{array}{ll}
    \textbf{Mask}(f_T^i,~\hat{e_i}=1), &\text{training} \\
    \textbf{Mask}(f_T^i,~e_i> \beta), &\text{inference} \\
\end{array}
\right.
\end{aligned}
\label{eq:fa}
\end{equation}
To achieve point cloud completion, we perform tasks at multiple resolutions and calculate the average loss for each voxel at that level. Then this average loss is summed and divided by $N-1$. The smooth L1 loss function is used to supervise the completion training process:
\begin{equation}
    \begin{aligned}
    \mathcal{L}_T = \frac{1}{N-1}   \sum_{i=1}^{N-1}\frac{1}{num_i}\textbf{SmoothL1}(e_i,\hat{e_i})
    \end{aligned}
\end{equation}
where $num_i$ is the number of generated voxels in level $i$.
The loss for training the whole framework is defined as
\begin{equation}
    \begin{aligned}
    \mathcal{L} = \mathcal{L}_D + \alpha\mathcal{L}_T
    \end{aligned}
\end{equation}
where $\mathcal{L}_D$ represents the loss for the detection pipeline and $\alpha=3.0$.
Since the detection network operates before the completion network, we can extract the detection results before initiating the completion network.
Additionally, when speed is not highly prioritized, an additional detector that adopts a completion point cloud as input can be implemented to achieve higher detection accuracy.

\begin{figure}[!t]
\includegraphics[width=0.99\linewidth]{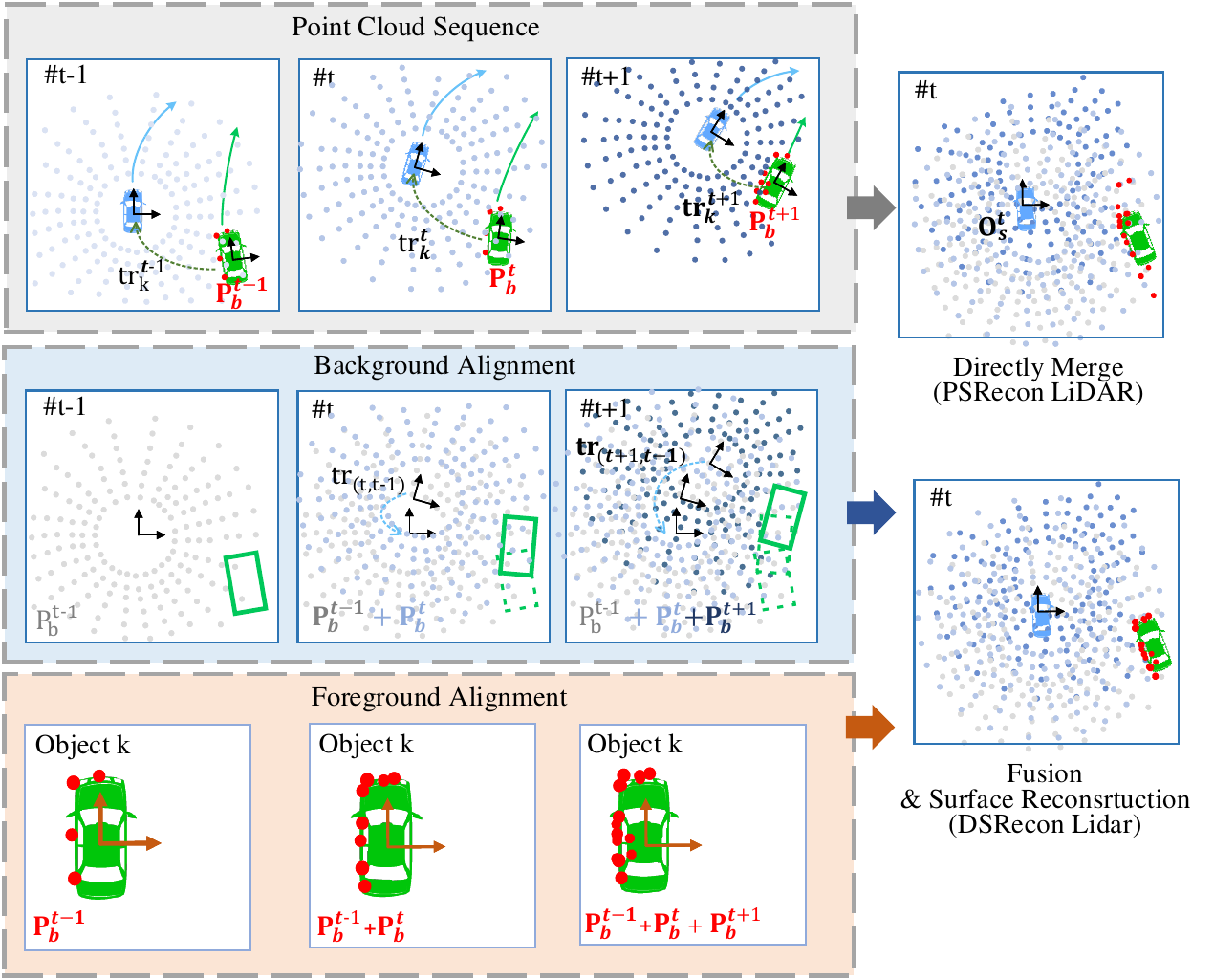}
\centering
\caption{\textbf{Comparison of the DM data and DSRecon data}. We show the alignment of three successive point cloud frames as an example. The top row shows the original point cloud sequence. Similar to DM, the background point clouds are aligned with $o_s^t$ (the middle row). The target point clouds are aligned in their coordinates  $o_k^t$ (the bottom row). The {\color{blue}blue} car represents the ego-vehicle, the {\color{green}green} car indicates a moving object in this sequence, and the {\color{green}green} boxes represent the 3D box annotations of this object.
}
\label{fig:sdf}
\end{figure}
\subsection{Dynamic-Static Reconstruction for Data Generation}
\label{sec:gt}
Dense ground truth point clouds are crucial for training the completion network. Traditional methods, such as those employed in SVCN~\cite{yi2021complete}, may produce inaccurate point clouds when dynamic objects are involved, often causing noise in distant areas.
To solve the challenge, we propose the \textit{Dynamic-Static Reconstruction} (DSRecon) module to enhance the foreground and background and perform reconstruction in sparse regions.
For ease of understanding, we define a typical point cloud sequence containing $T$ frames. The point cloud for a single LiDAR sweep is denoted as $P^t$, where $t\in \{1,2,...T\}$. Each $P^t$ contains the background point cloud $P_b$, and the point clouds of $K$ objects $P_{k}^t$, where $k\in \{1,2,...K\}$.  
\textbf{Dynamic Foreground.}
We create a dense point set $P_{k}$, by gathering point clouds from different frames $T$ of the same object to its center. This alignment relies on the 3D box annotations of each frame. We define the $(x,y,z)$ transformation matrix from the LiDAR sensor to the center of the object $k$ as $\mathbf{tr}_k^t$.
\begin{equation}
\label{eq:fg}
\begin{aligned}
P_{k} = \{P_{k}^t\cdot inv(\mathbf{tr}_k^t)|t=1,2,...T\}
\end{aligned}
\end{equation}
\textbf{Static Background.}
The alignment process considers only the foreground point clouds in the sequence, without accounting for the background blending effect. We argue that acknowledging diverse backgrounds is crucial in 3D object detection. 
The static background $P_b^t$ of each sweep is obtained by removing $P_{k}^t$ from the sweep point cloud. Then, we merge all the background point clouds in the sequence with the transformation metric from sweep $t$ to sweep $1$, $\mathbf{tr}_{(t,1)}$.
\begin{equation}
\label{eq:fg2}
\begin{aligned}
P_{b} = \{P_b^t\cdot \mathbf{tr}_{(t,1)}~|~t=1,2,...T\}
\end{aligned}
\end{equation}

\textbf{DSRecon Data.}
We transform the dense foreground point clouds $P_{k}$ back to their respective positions in each frame using $\mathbf{tr}_k^t$, denoted  as $R_{k}^t$ and background point cloud  $R_{b}$ to the sweep $t$. For the background and object point clouds aligned with more than 200 points, we perform surface reconstruction using NKSR~\cite{huang2023neural}, followed by uniform resampling of the point clouds. Subsequently, we combine the fore/background to create $\text{DSRecon}^{t}$ point cloud data for sweep $t$.
\begin{equation}
\label{eq:fl}
\begin{aligned}
\text{DSRecon}^{t} = \{R_{b}^t\cup \{R_{k}^t~|~k=1,2...K\}\} ~~~\\
R_{b}^t=\{\text{NKSR-Resample}(P_{b})\cdot\mathbf{tr}_{(1,t)}~|~t=1,2...T\} \\
R_{k}^t=\text{NKSR-Resample}(P_{k}^t)\cdot\mathbf{tr}_k^t~~~~~~~~~~~\\
\end{aligned}
\end{equation}

\begin{table*}[ht!]
\centering
\caption{Evaluation results for adapting TransBridge to different detectors on nuScenes validation split.
}
\resizebox{0.95\textwidth}{!}{
\begin{tabular}{c|ccc|llccccc}
\thickhline
\rowcolor{mygray}
\multicolumn{1}{c}{Backbone}                          & Encoder & Decoder & Stage-2 & \textbf{~mAP}$\uparrow$ & \textbf{~NDS}$\uparrow$& mATE$\downarrow$ & mASE$\downarrow$ & mAOE$\downarrow$ & mAVE$\downarrow$ & mAAE$\downarrow$  \\ \hline
\hline\rule{0pt}{9pt}
\multirow{3}{*}{Second(2018)~\cite{yan2018second}}
&\checkmark & & & 50.59 & 62.29 & 31.15 & 25.51 & 26.64 & 26.26 & 20.46 \\
&\checkmark &\checkmark &
&51.94
&63.42
& \textbf{30.89}&25.61& \textbf{23.25}& \textbf{25.89}&19.80\\
&\checkmark &\checkmark &\checkmark & \textbf{56.22}  &\textbf{64.77} & 31.55 &  \textbf{25.05} & 28.83 & 30.10 &  \textbf{19.31}\\ \hline
    \rule{0pt}{9pt}
\multirow{3}{*}{PointPillar(2019)~\cite{lang2019pointpillars}}                                                                  &\checkmark & & &  44.63 & 58.23 & 33.87 & 26.00 & 32.07  & 28.74 & 20.15     \\
&\checkmark &\checkmark &
&45.78
&58.08
&\textbf{33.49}&25.89&\textbf{31.35}&\textbf{27.95}&20.77 \\
&\checkmark &\checkmark &\checkmark & \textbf{47.79} & \textbf{59.39} & 35.54 & \textbf{25.59} & 32.81 & 31.93 & \textbf{19.17} \\\hline
\rule{0pt}{9pt}
\multirow{3}{*}{\begin{tabular}[c]{@{}c@{}}Pillar-Based\\ Centerpoint(2021)~\cite{yin2021center}\end{tabular}}
&\checkmark & & & 50.03 & 60.70 & 31.13 & 26.04& 42.92 &  \textbf{23.90} & 19.14 \\
&\checkmark &\checkmark &
&51.48
&60.72
& \textbf{30.14}&26.27&43.25&31.50& \textbf{19.01}\\
&\checkmark &\checkmark &\checkmark & \textbf{52.93} & \textbf{62.14}  & 31.74 &  \textbf{25.46} &  \textbf{33.60} & 33.43 & 19.08\\ \hline\rule{0pt}{9pt}
\multirow{3}{*}{\begin{tabular}[c]{@{}c@{}}Voxel-Based\\ Centerpoint(2021)~\cite{yin2021center}\end{tabular}}
&\checkmark & & &
56.03 & 64.54 & 30.11 & 25.55 & 38.28 & \textbf{21.94} & 18.87 \\ 
&\checkmark &\checkmark &
&56.97
&65.21
& \textbf{29.40}& \textbf{25.47}& \textbf{37.01}& 22.69& \textbf{18.13} \\
&\checkmark &\checkmark &\checkmark &\textbf{57.88}&	\textbf{65.34}&	30.36&	25.67&	38.44&	22.03&	19.46
 \\ \hline
\multirow{3}{*}{\begin{tabular}[c]{@{}c@{}}VoxelNext(2023)~\cite{chen2023voxelnext}\end{tabular}}
\rule{0pt}{9pt}
&\checkmark & &  &60.53&66.65&30.11&25.23&29.35&27.31&\textbf{18.55} \\
&\checkmark &\checkmark & &61.19&67.76 & \textbf{29.88}	& \textbf{25.12}& \textbf{27.45	}& \textbf{27.09} & 18.74\\
&\checkmark &\checkmark &\checkmark &\textbf{62.21}& \textbf{67.97}& 30.24& 25.37& 29.51& 27.44&	18.75\\ 
\hline
\multirow{3}{*}{\begin{tabular}[c]{@{}c@{}}HEDNet(2023)~\cite{zhang2023hednet}\end{tabular}}
\rule{0pt}{9pt}
&\checkmark & &  &66.82&71.05&27.38&25.07&26.69&25.80&\textbf{18.68} \\
&\checkmark &\checkmark & & 67.78 & 71.89 &\textbf{26.24} &\textbf{24.36} &\textbf{25.82}&\textbf{24.71} & 18.88\\
&\checkmark &\checkmark &\checkmark &\textbf{68.04} &\textbf{71.93}& 27.01 &	24.58&	25.91&	24.75&	19.36
\\ 
\hline
\end{tabular}
}
\label{table:nus}
\end{table*}

\section{Experiments}
\subsection{Implementation Details}
\subsubsection{Feature Extraction from Detection}
We implement TransBridge blocks in a 10-frame voxel-based Centerpoint on the nuScenes dataset \cite{nuscenes2019}. We use \textit{voxel size}=[0.1,0.1,0.2] meters, where 0.2m represents the height of each voxel. Thus, the size of the feature map is [1024,1024,40].
Each voxel contains the average point features (x, y, z, intensity, timestamp) within it.
The pipeline begins with submanifold convolution applied to the voxel features, which yields a 16-channel feature map $f_D^1$.
This step is followed by a series of sparse convolutions with a stride of 2 and submanifold convolutions with a stride of 1, which serve as detection encoders. 
These encoders produce detection feature maps $f_D^2, f_D^3, ..., f_D^5$ with channel counts of (16, 32, 64, 64, 128).
The final feature map $f_D^5$, is fed into the detection head for 3D box prediction.

\subsubsection{Feature Up-Sampling via TransBridge}
For the lowest resolution level, we simplify the TransBridge block to $f_T^{5} = \mathbf{MLP}(f_D^{5})$. This block contains 128 channels and produces a feature map with dimensions (64, 64, 1). The other TransBridge blocks employ a transformer structure. The completion feature maps in these blocks have channel counts of (128, 160, 64, 32, 16).
It is worth noting that the TransBridge is designed to construct Bird's Eye View (BEV) existence maps for BEV-based methods like PointPillar.

\subsubsection{Multi-Resolution Ground Truth Generation}

The voxelization process across all resolution levels is computationally demanding. To manage this, we compute max-pooling for the larger resolution levels using the same kernels as those used in the down-sampling scales: ([2,2,2], [2,2,2], [2,2,2], [1,1,5]). This approach enables us to obtain all the voxel existence scores, $e^i$, from a single voxelization operation of $e^1$.

\subsubsection{Sparsity Control Module (SCM)}
During the inference, TransBridge blocks up-sample lower-resolution feature maps to create higher-resolution ones, significantly increasing the number of new voxels.
 For example, each voxel in the last level generates 5 voxels in the next feature map with a stride of [1,1,5], and 8 voxels with a stride of 2.
This upsampling process will increase memory requirements. To mitigate this, the SCM adjusts the number of voxels progressing to the next level.
Throughout training, we maintain a ratio of \textit{empty voxels} to \textit{all generated voxels} below 0.75, balancing positive and negative samples.
Only those voxels with a matching counterpart in ground truth, $e^i$, are preserved to the next level. During inference, voxel preservation is determined using a threshold value of $\beta=0.7$ to maintain sparsity.

\begin{figure*}
\includegraphics[width=0.99 \linewidth]{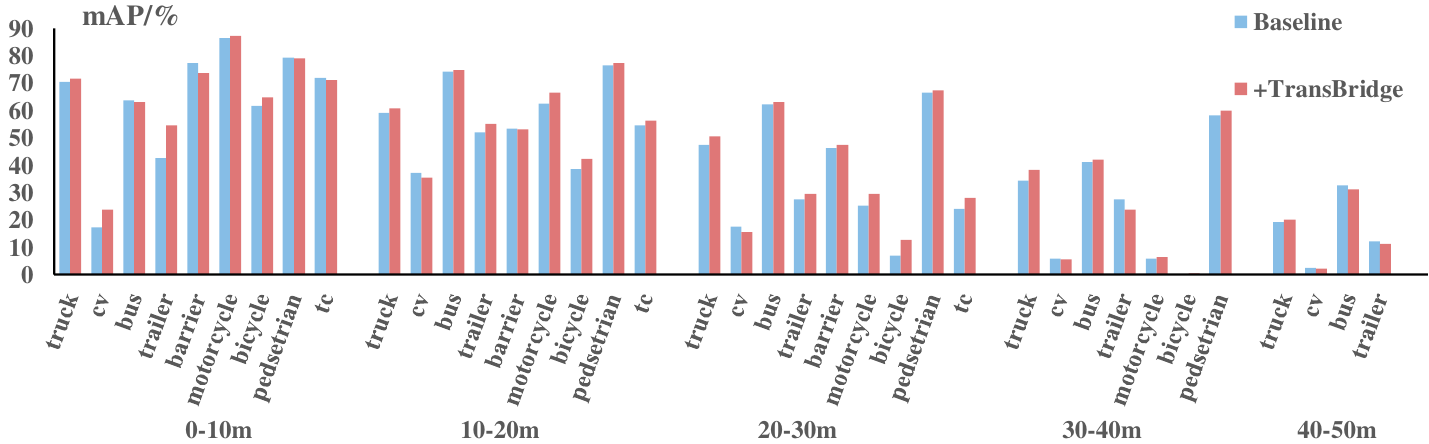}
\centering
\caption
{
\textbf{Statistical results of the object detection performance along different distances for each category.} TransBridge improves the performance between other classes, especially in distant regions. The `cv' and `tc' are short for `\textit{construction vehicle}' and  `\textit{traffic cone},' respectively. 
Our approach demonstrates superior performance compared to other methods in most categories, particularly showing significant improvements in close-to-medium range detections. This emphasizes the effectiveness of our approach in achieving better results across various object categories.} 
\label{fig:supp_dist}

\vspace{5mm}
\includegraphics[width=0.99 \linewidth]{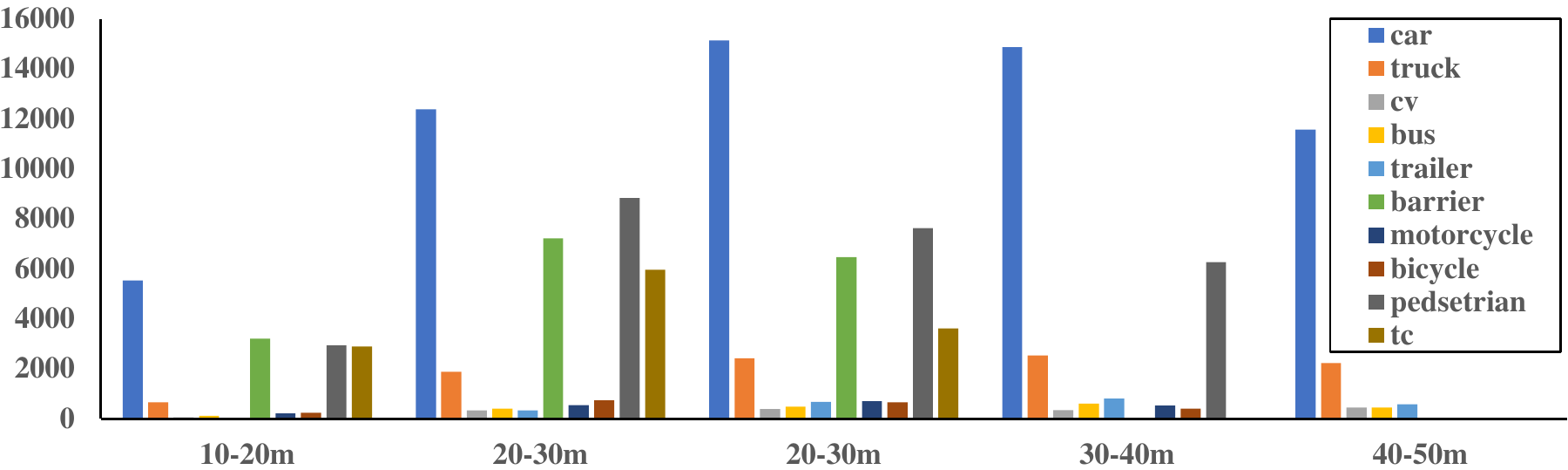}
\centering
\caption
{
\textbf{The number of objects in the different distances on nuScenes dataset}. The `cv' and `tc' are short for `\textit{construction vehicle}' and  `\textit{traffic cone},' respectively. We can see that `car', `truck', `barrier', `pedestrian', and `traffic cone' are the main categories, and the rest are long-tailed categories.
The figure highlights that certain categories, such as construction vehicles (`cv') and barriers, fall into long-tail categories with limited instances across different distances, particularly in the far range. This distribution of instances sheds light on the performance discrepancies observed for these specific categories.} 
\label{fig:supp_num}
\end{figure*}

\subsection{Datasets and Metrics}
\textbf{nuScenes Dataset}~\cite{nuscenes2019} consists of 1,000 driving scenes in ten labeled categories, each lasting 20 seconds and annotated at a frequency of 2 Hz. The annotated ones are called `samples' and the others are called  `sweeps'.
 The nuScenes Detection Metrics utilize two primary metrics: the \textit{mean Average Precision} (mAP) and the \textit{NuScenes Detection Score} (NDS). The mAP evaluates based on the center distance threshold, and NDS is a weighted sum of various \textit{True Positive} (TP) metrics. In our experiments, we use nine sweeps and one sample as input, generating DSRecon point cloud data by fusing around 40 samples from the full sequence.
\begin{equation}
\begin{aligned}
NDS = \frac{1}{10}\left[5mAP + \sum_{TP}(1-min(1,TP))\right]
\end{aligned}
\label{eq:nds}
\end{equation}

\textbf{WOD Detection.}
We also utilize the Waymo Open Dataset (WOD)~\cite{sun2020scalability}, which comprises 798 training and 202 validation sequences focused on vehicles. WOD employs a 64-lane LiDAR sensor for denser scans and provides annotations every 0.1 seconds. Its evaluation metrics include 3D bounding box mAP and mAP weighted by heading accuracy (mAPH), based on an IoU threshold of 0.7 for vehicles. 
We use a fifth of the annotated frames for training and conduct experiments on multiframe sequences spanning [-0.3, 0s].

\subsection{Evaluation Results}

To demonstrate the effectiveness of our methods, we apply them to four 
widely used LiDAR-based 3D detectors: SECOND~\cite{yan2018second}, PointPillars~\cite{lang2019pointpillars}, Centerpoint~\cite{yin2021center}, and VoxelNext~\cite{chen2023voxelnext}.
We conducted experiments using three different strategies to comprehensively evaluate performance with different baselines.  
{To ensure reproducibility and accuracy, we utilized the official configuration files obtained from the GitHub repositories of the nuScenes and WOD datasets for the implementation of all referenced state-of-the-art algorithms. Our method is trained from scratch using the same training settings as the compared method instead of finetuning based on them, which ensures fairness.}
1) We present the baseline method, in which only the detection encoder is trained (denoted as \textbf{Encoder}); 2) In the second line, we adapt the TransBridge completion decoder to achieve joint learning (denoted as \textbf{Decoder}); 3) In 3rd line, we collect the completion point cloud and refine the detection encoder as a second stage to achieve higher detection performance (denoted as \textbf{ Stage-2}), which will be detailed in Section~\ref{sub:cas}.
 
The joint manner exceeds all 5 baselines, and an improvement of 1.0 in the mAP and NDS metrics is obtained with end-to-end training. In addition, the two-stage manner achieves the best performance on mAP and NDS.
However, we find these manners may fail on sub-metrics. 
We attribute this to the fact that the completion point cloud blurred the precise location of the object, thereby slightly increasing the positioning error, 
which is a direction we aim to investigate further. 

\begin{table}[!t]
\centering
\caption{the computational cost of different strategies.
}
\resizebox{0.44\textwidth}{!}{
\begin{tabular}{c|c|c}
\rowcolor{mygray}
\thickhline
Method(Centerpoint) & Inference Time & GPU Memory  \\\hline\hline
\rule{0pt}{9pt}

Baseline &107.4ms &4917MB \\   
One-Stage &107.9ms &4966MB\\
Two-Stage &454.1ms &13566MB\\\hline\bottomrule
\end{tabular}
}
\label{table:time}
\end{table}

\subsection{Cascade Framework}
\label{sub:cas}

Using a two-stage strategy, we observe an increase of up to 5.78 points in mAP for the SECOND model~\cite{yan2018second}.
~\tabref{table:time} below provides a comparison of memory usage and inference time between the baseline detector using a one-stage network and the cascade network. We observe that Lines 1 and 2 share similar inference times as mentioned before.
The introduction of the second-stage network necessitates completing the first-stage feature extraction and point cloud completion before executing the second-stage object detection. As a result, this approach incurs significant computational costs (3rd line). 
\begin{figure*}[ht!]
\includegraphics[width=0.99 \linewidth]{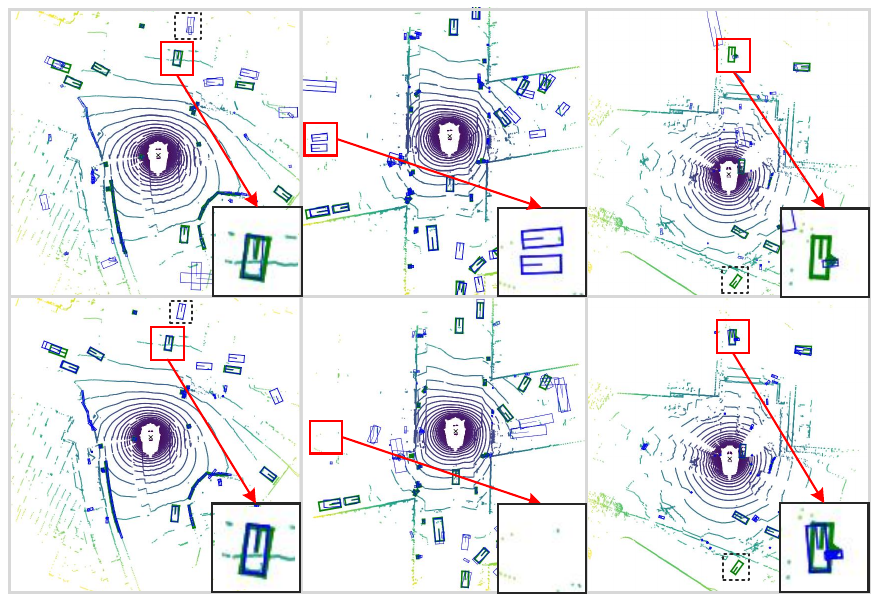}
\centering
\put(-440,-10){ \textbf{Accurate}}
\put(-270,-10){ \textbf{Clean}}
\put(-120,-10){  \textbf{Comprehensive}}
\caption{\textbf{The comparison of the qualitative results of our detection without and with TransBridge on nuScenes.} The sub-figures in the first row show the results of the baseline detector, and the bottom row provides the results of TransBridge detector.
The ground truth boxes are shown in {\color{green}green}, and the predicted boxes are colored in {\color{blue}blue}
The detection results are more accurate (left), cleaner (middle), and more comprehensive (right). 
}
\label{fig:visu_det}
\end{figure*}

\subsection{Generalization Capabilities}
\textbf{Generalization on specific categories.}
To demonstrate the generalization capabilities of our proposed method across a wide range of object categories, \figref{fig:supp_dist} visually represents the detection accuracy of mAP, for various object classes on the nuScenes validation split. The performance of the baseline is represented by blue markers, while the results of our detector enhanced with the TransBridge are indicated by red markers.
Our approach outperforms the baseline in most categories, demonstrating its superiority in handling diverse object types. This is particularly evident for close-to-medium-range detections, underscoring the enhanced robustness of our framework. {Additionally, the $1.5$ mAP improvement for cars beyond 50 meters enhances safe reaction time by approximately 200 milliseconds in highway driving situations.}
However, some categories exhibit weaker performance, notably construction vehicles.
To analyze the factors that contribute to these performance variances, \figref{fig:supp_num} offers a detailed analysis of the distribution of annotated bounding boxes across distances for each object class. The analysis shows that construction cars belong to a subset of long-tail categories, characterized by a scarcity of instances at various distances, particularly in the far range. 

\begin{table}[!t]
\centering
\caption{Evaluation results on WOD dataset validation split.
}
\resizebox{0.44\textwidth}{!}{
\begin{tabular}{c|cc|cc}
\rowcolor{mygray}
\thickhline
\multirow{2}{*}
  & \multicolumn{2}{c} {L1} & \multicolumn{2}{c} {L2} \\\cline{2-5}
\rowcolor{mygray}\multirow{-2}{*}
{Method} & 3D AP$\uparrow$ & 3D APH  $\uparrow$~~ & 3D AP  $\uparrow $ & 3D APH$\uparrow$\\
               \hline\hline
\rule{0pt}{9pt}
Centerpoint~\cite{yin2021center}
& 74.91&74.38& 67.10&66.67  \\    
+TransBridge
& \textbf{75.75} &  \textbf{75.22}&  \textbf{67.98} & \textbf{ 67.49} \\
\hline
\rule{0pt}{9pt}
MPPNet~\cite{chen2022mppnet}
& 80.21& 79.73 & 72.53& 72.09 \\
+TransBridge
&  \textbf{80.94} &  \textbf{80.51} &  \textbf{73.26} & \textbf{72.82} \\ \hline
\end{tabular}
}
\label{table:waymo}
\end{table}

\textbf{Generalization on WOD dataset.}
Our experiments with the WOD dataset~\cite{sun2020scalability} use 4-frame Centerpoint~\cite{yin2021center} and MPPNet~\cite{chen2022mppnet} as baselines.
The DSRecon point cloud data is generated using approximately 40 keyframes, similar to nuScenes. 
In~\tabref{table:waymo}, TransBridge improves the performance of both detectors by about 1 point across both difficulty levels.

\subsection{Ablation Study}
\begin{figure}[ht!]
\includegraphics[width= 1.00 \linewidth]{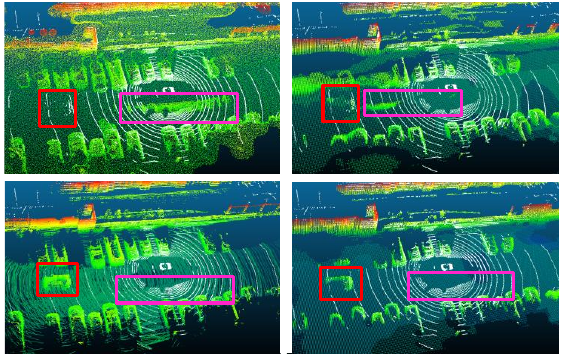}
\centering
\put(-250,85){ {\color{white}DM Data}}
\put(-120,85){ {\color{white}DM \& Completion}}
\put(-250,5){ {\color{white}DSRecon Data}}
\put(-120,5){ {\color{white}DSRecon \& Completion}}
\caption{\textbf{The qualitative results of 3D point cloud completion with DM data and DSRecon data on nuScenes.}
The two sub-figures in the left column are DM point cloud data and DSRecon data, and the two sub-figures in the right column are predicted dense point clouds. }
\label{fig:visu_comp}
\end{figure}
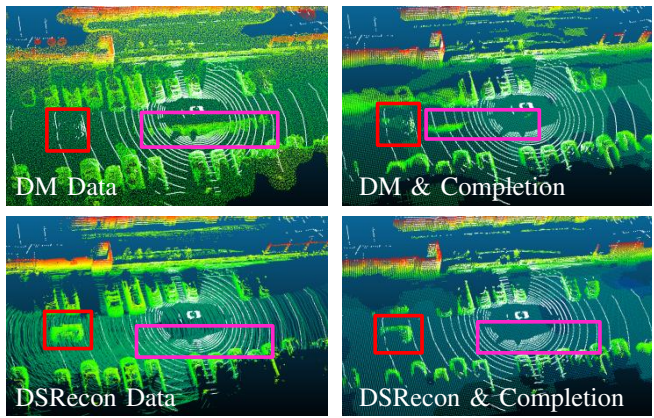

We conduct a series of experiments to evaluate each component of our framework, using a quarter of the training data from the nuScene dataset~\cite{nuscenes2019} for quick evaluation. The results from the complete validation split reveal significant insights.

\noindent{\textbf{Ablation on Data Generation.}}

Our experimental results, summarized in \tabref{table:ablation_study_data}, explore the impact of different point cloud data types on learning detection and completion.
Directly Merged (DM) data negatively affects detector performance, highlighting the critical need for precise handling of trailing smears.
However, aligning the foreground-merged (FM) and background-merged (BM) data separately shows significant improvements in mAP, emphasizing the importance of foreground and background completion in enhancing detection accuracy.
The combination of FM and BM data achieves the highest mAP of 53.51. Furthermore, employing the refined Surface Reconstruction (SR) from NKSR~\cite{huang2023neural} surpasses all other strategies, boosting the baseline detector's mAP to 53.98.

\begin{table}[!t]
\centering
\caption{Evaluation results for Data Generation Strategies.    }
    \resizebox{0.35\textwidth}{!}{
    \begin{tabular}{cccc|ll}
    \thickhline
    \rowcolor{mygray}
    \rule{0pt}{9pt}
DM\cite{yi2021complete} &FM &BM & SR & \textbf{mAP}$\uparrow$  & \textbf{NDS}$\uparrow$ \\
\hline\hline\rule{0pt}{9pt}
&&&& 52.52 & 61.26       \\
\checkmark&&&& 52.16 $\searrow$   &60.79$\searrow$         \\
&\checkmark&&& 53.42&61.86    \\
&&\checkmark&& 53.28&61.62   \\
&\checkmark&\checkmark&&53.51&62.17\\
&\checkmark&\checkmark&\checkmark&\textbf{53.98}    &\textbf{62.20}     \\ \hline
    \end{tabular}
    }
    \label{table:ablation_study_data}
\end{table}

\noindent{\textbf{Ablation on Network Choice.}}
As detailed in Table \ref{table:ablation_study_net}, we evaluate various network components within the TransBridge framework. These experiments are categorized into four sections:
\textit{1) Channel-Cut Structure\cite{yi2021complete}:} Minor performance variations indicate that this approach may not be optimal for guiding the detection network.
\textit{2) FoldingNet-like Structures \cite{yang2018foldingnet}:} Incorporating these structures, along with normal (FN) and biased (FB) positional embeddings, leads to decreased performance, suggesting their incompatibility with our framework's requirements.
\textit{3) TransBridge Structures (Ours)}: 
The adoption of the Up-Sample Bridge (UB) shows a clear improvement in mAP, indicating its effectiveness.
Additionally, the single Interpreting Bridge (IB) further enhances mAP, underscoring the need to recognize differences between detection and completion features. The TransBridge module achieves the highest mAP of 53.98, validating its comprehensive effectiveness.
{\textit{4) Dropping Levels (Ours)}:
We analyze the impact of removing the final reconstruction levels in the TransBridge network. Our results show that removing the last two levels (Drop 1st, 2nd) of high-resolution reconstruction during training had virtually no effect on object detection performance. 
However, after removing up to three levels of reconstruction results (Drop 3rd), the benefits of reconstruction for detection disappeared, suggesting that overly coarse reconstruction does not contribute to the detection network.}

\begin{table}[!t]
\centering
\caption{Evaluation results for different network structures.
}
    \resizebox{0.47\textwidth}{!}{
    \begin{tabular}{c|c|cc|cc|ll}
    \thickhline
    \rowcolor{mygray}
    \rule{0pt}{9pt}
Decoder &CC & FN & FB & UB & IB   & \textbf{~mAP} $\uparrow$ & \textbf{~NDS}$\uparrow$ \\
\hline\hline
\multirow{2}{*}{Baseline}&&&&&& 52.52 & 61.26     \\
&\checkmark&&&&&52.57    &61.38     \\\hline
\multirow{2}{*}{FoldingNet~\cite{yang2018foldingnet}}&&\checkmark&&&&51.98$\searrow$ & 60.92$\searrow$   \\
&&&\checkmark&&&48.89$\searrow$ &57.88$\searrow$          \\ \hline
\multirow{3}{*}{Transformer}&&&&\checkmark&&53.16    &61.56        \\
&&&&&\checkmark&53.64    &62.07        \\
&&&&\checkmark&\checkmark&\textbf{53.98}    &\textbf{62.20}\\ \hline
{Drop 1st }      &&&&\checkmark&\checkmark&53.87     &62.19          \\
{+Drop 2nd}     &&&&\checkmark&\checkmark&53.82     &62.08         \\ 
{+Drop 3rd }     &&&&\checkmark&\checkmark&53.25     &61.83      \\\hline
\end{tabular}
    }
\label{table:ablation_study_net}
\end{table}

\subsection{Qualitative Results}
\noindent\textbf{3D Object Detection.} In ~\figref{fig:visu_det}, we visualize the output of the baseline detector (top) and our TransBridge-enhanced detector  (bottom)  on the nuScenes dataset.
We render a single LiDAR sweep here to visualize the bounding boxes. The ground truth boxes are shown in {\color{green}green}, and the predicted boxes are colored in {\color{blue}blue}.
Key observations include: 1)
\textit{1) Improved Accuracy}: TransBridge significantly enhances the accuracy of 3D box predictions, as demonstrated by the more accurate orientation of vehicles and higher Intersection over Union (IoU) scores.
\textit{2) Reduced False Positives}: Particularly in distant regions, TransBridge effectively lowers the rate of false positives.
\textit{3) Enhanced Distant Object Detection}: The network's ability to recognize distant objects is markedly improved with TransBridge, as shown in the visualizations.

\noindent\textbf{3D Point Cloud Completion.} The comparison between the DM (top left) and DSRecon (bottom left) data in \figref{fig:visu_comp} demonstrates the effectiveness of DSRecon in eliminating trailing smears and accurately reconstructing vehicles.
DSRecon data eliminates trailing smear, as shown in the \textcolor[RGB]{206,108,166}{purple} boxes. Furthermore, the vehicle in the {\color{red}red} boxes is correctly reconstructed, which is not the case with the DM data.
This indicates that DSRecon data is more suitable for scenarios involving moving objects. 

{
\noindent\textbf{Limitations:} TransBridge's performance deteriorates for objects with extreme sparsity  (e.g., $<5$ points beyond 80m). Refer to the dashed boxes in \figref{fig:visu_det}. The box in the left subfigure represents a false positive (blue boxes without green ground truth annotations), while the box in the right subfigure denotes a false negative (green ground truth annotations without corresponding blue boxes).
Additionally, the point cloud completion network nearly doubles the consumption of training. We plan to conduct an investigation of integrating temporal fusion of multiple point cloud frames and adopt a pre-training strategy to mitigate the limitations in future work.
}

\vspace{3mm}
\section{Conclusion}
This paper introduces a novel framework that enables joint training of end-to-end 3D object detection and point cloud completion. 
{It significantly improves the performance of various LiDAR perception systems in ITS, particularly in distant regions with sparse data.}
The proposed TransBridge incorporates two key components: the specially designed Up-Sample Bridge and the Interpreting Bridge. The Up-Sample Bridge extracts informative features that compensate for information loss during the channel-cut up-sampling process. The Interpreting Bridge facilitates efficient feature delivery between the detection and completion networks.
Additionally, the Sparsity Control Module (SCM) ensures that the density of the point clouds is suitable for both the training and inference stages. To generate high-quality training data, the DSRecon module produces dense point clouds without noise.
Furthermore, TransBridge has been applied to several state-of-the-art 3D object detectors, leading to significant performance improvements on the nuScenes and WOD datasets.

{\small \normalem
\bibliographystyle{IEEEtran}
\bibliography{egbib}
}

\vfill

\end{document}